\pdfoutput=1

\documentclass[11pt]{article}
\PassOptionsToPackage{dvipsnames}{xcolor}
\usepackage[]{acl}
\usepackage{array, makecell} 
\usepackage{times}
\usepackage{latexsym}
\usepackage{graphicx}
\usepackage{amsfonts}
\usepackage{amsmath}
\usepackage{amssymb}
\usepackage{booktabs} 
\usepackage{multirow}
\usepackage{cleveref}
\usepackage{xspace}
\usepackage{enumitem}

\crefformat{section}{\S#2#1#3}
\crefmultiformat{section}{\S\S#2#1#3}{ and~#2#1#3}{, #2#1#3}{, and~#2#1#3}

\usepackage[T1]{fontenc}

\usepackage[utf8]{inputenc}

\usepackage{microtype}
\usepackage[most]{tcolorbox}
\colorlet{LightLavender}{lightgray!20!}
\tcbset{on line, 
        boxsep=4pt, left=0pt,right=0pt,top=0pt,bottom=0pt,
        colframe=white,colback=LightLavender,  
        highlight math style={enhanced}
        }

\newcommand{\blue}[1]{\textcolor{blue}{\texttt{#1}}}
\newcommand{\black}[1]{\textcolor{black}{\texttt{#1}}}

\newcommand{\afterprompt}[1]{
&             \blue{Document:} \black{[D1]} \blue{Document:} \black{[D2]} \blue{, ...,} \\
&             \blue{#1} \\
&             \blue{Question:} \\
}
\newcommand{\beforeprompt}[1]{
&             \blue{#1} \\
&             \blue{Document:} \black{[D1]} \blue{Document:} \black{[D2]} \blue{, ...,} \\
&             \blue{Question:} \\
}

\usepackage{microtype}

\newcommand{\ignore}[1]{}
\newcommand{\ssymbol}[1]{^{\@fnsymbol{#1}}}

\newcommand{\model}{\textsc{PromptRank}\xspace}
\newcommand{\gptxl}{GPT2-XL}
\newcommand{\txl}{T5-XL}
\newcommand{\tlarge}{T5-Large}
\newcommand{\tbase}{T5-Base}

\newcommand{\cutsectionup}{\vspace*{-0.00in}}
\newcommand{\cutsectiondown}{\vspace*{-0.00in}}

\newcommand{\cutsubsectionup}{\vspace*{-0.00in}}
\newcommand{\cutsubsectiondown}{\vspace*{-0.002in}}

\newcommand{\cutparagraphup}{\vspace*{-0.00in}}

\newcommand{\cutcaptionup}{\vspace*{-0.00in}}
\newcommand{\cutcaptiondown}{\vspace*{-0.0in}}
\newcommand{\cutcaptiondownvery}{\vspace*{-0.0in}}

\title{Few-shot Reranking for Multi-hop QA via Language Model Prompting}

\author{Muhammad Khalifa\thanks{$\,\,\,$Correspondence to \tt{khalifam@umich.edu}}\hspace{3pt}, Lajanugen Logeswaran$^\dagger$, \textbf{Moontae Lee}$^{\dagger\ddagger}$, \\ 
\textbf{Honglak Lee}$^{*\dagger}$\textbf{,} \textbf{Lu Wang}$^*$ \\
University of Michigan$^*$, LG AI Research$^\dagger$, University of Illinois at Chicago$^\ddagger$
}

\date{}

\begin{document}
\maketitle

\begin{abstract}
We study few-shot reranking for multi-hop QA with open-domain questions. 
To alleviate the need for a large number of labeled question-document pairs for retriever training, we propose \textbf{\model}, which relies on language model prompting for multi-hop path reranking. \model first constructs an instruction-based prompt that includes a candidate document path and then computes the relevance score between a given question and the path based on the conditional likelihood of the question given the path prompt according to a language model.
\model yields strong retrieval performance on HotpotQA with only 128 training examples compared to state-of-the-art methods trained on thousands of examples --- 73.6 recall@$10$ by \model vs. 77.8 by PathRetriever \cite{Asai2019} and 77.5 by multi-hop dense retrieval \citep{mdr2021}.\footnote{Code available at 
\url{https://github.com/mukhal/PromptRank}.
}


\end{abstract}

\cutsectionup
\section{Introduction}
\cutsectiondown

Many information-seeking queries are in the form of multi-hop questions. For instance, to answer the question \textit{``What 1988 Christmas comedy film did Brian-Doyle Murray star in?''}, we need to \textbf{(i)} search for movies starring Brian Murray, then \textbf{(ii)} identify which of them were released in 1988 during Christmas.
Evidence required to answer such questions is often dispersed in different documents, requiring sequential, multi-step reasoning to reach the answer~\cite{perez20}, typically referred to as multi-hop question answering (MQA).

\begin{figure}[t!]
    \centering
    \includegraphics[width=0.95\linewidth]{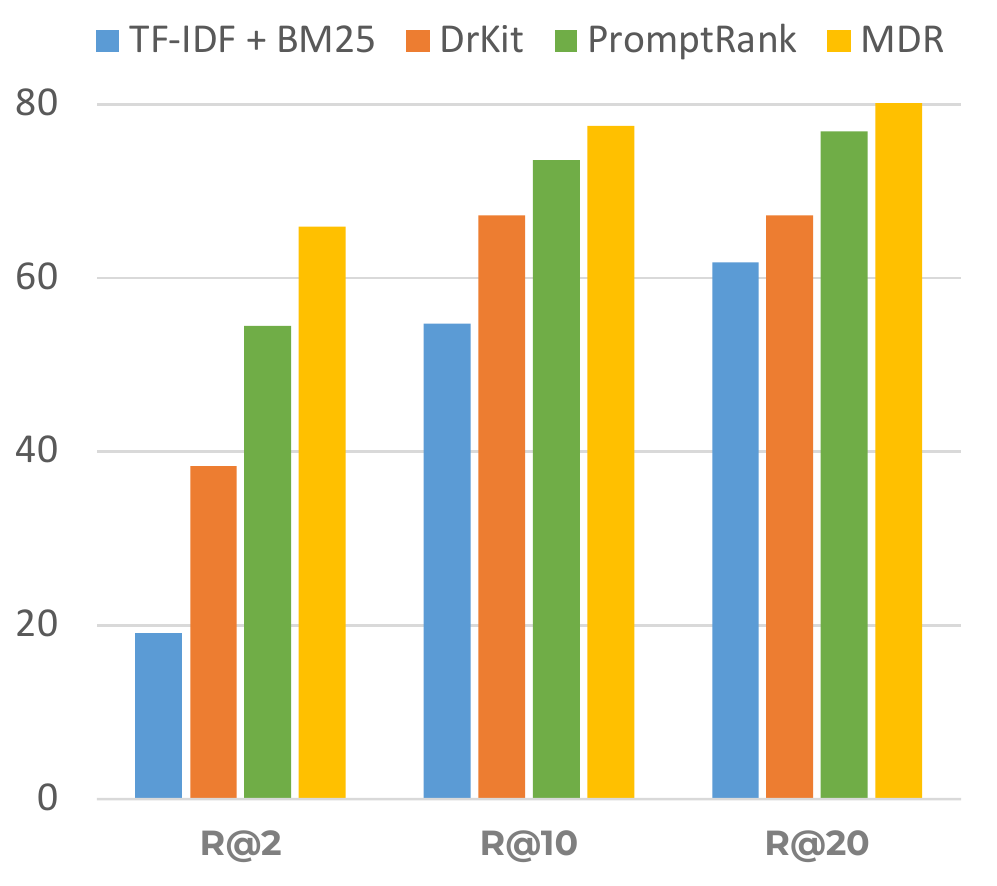}
    \cutcaptionup
    \caption{
    Multi-hop retrieval recall@$k$ on HotpotQA. \model, using only 128 examples, outperforms DrKit \citep{drkit2020} and performs closely to Multi-hop Dense Retrieval \citep{mdr2021}. Both fully supervised models are trained on \textasciitilde 90K examples. 
    }
    \cutcaptiondownvery
    \label{fig:teaser}
\end{figure}

\begin{figure*}
    \centering
    \includegraphics[width=15.8cm,height=5cm]{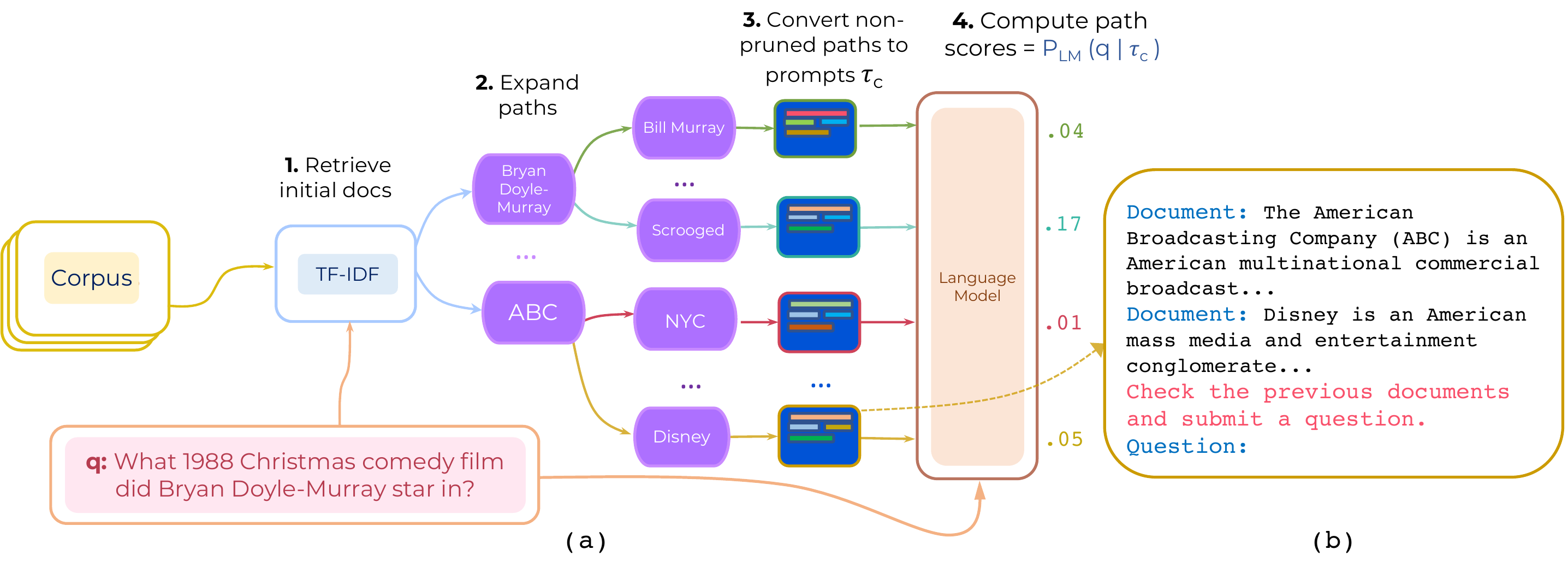}
    \cutcaptionup
    \caption{
    \footnotesize
    An overview of the full retrieval system. (a): Initial documents from TF-IDF are retrieved and expanded based on hyperlinks for $H$ times. \model converts each path into a prompt $\tau_c$ and scores through $P_{\text{LM}}(q | \tau_c)$ for a given question $q$ using a language model. 
    For simplicity, we omit intermediate scoring steps where paths of length $h < H$ are scored using the same fashion and only the top-scored ones are expanded. (b): A sample of what a 2-hop path prompt looks like. Prompts are constructed in terms of an \textcolor{WildStrawberry}{\textbf{instruction}} and the \textcolor{RoyalBlue}{\textbf{document path}}.}
    \label{fig:overview-prompt}
    \cutcaptiondown
\end{figure*}


Given a multi-hop question and a large document corpus, existing MQA systems largely follow a \textit{retrieve-then-read} pipeline, where a retriever module first identifies relevant documents from the corpus, and a reader module produces the answer based on the retrieved output \cite{Asai2019,li2021hopretriever,endtoend2021,qi2021answering}. 
The retriever module is trained to predict the ground-truth evidence document(s) given the question \cite{dpr2020,qi2021answering,}.
However, curating large datasets of question-document pairs is expensive, especially for low-resource languages or domains that require unique expertise (e.g., medical or legal documents), thus creating a \textit{bottleneck} for building QA pipelines ~\citep{ram-etal-2022-learning}. Moreover, resorting to heuristics for data labeling can lead to incorrect annotation \cite{distilling2021}. 
This difficulty is further exacerbated in the case of multi-hop questions, as they need to be annotated with multiple support documents. 

The majority of existing data-efficient retrieval and reranking methods are restricted to \textit{single-hop} QA, and it is unclear how to extend them to the \textit{multi-hop} setting. For instance, \newcite{ram-etal-2022-learning} proposed ``recurrent span retrieval'' to obtain \textit{psuedo} question-document pairs in an unsupervised way for single-hop QA. However, in the multi-hop case, it is less likely that we can retrieve recurrent spans from multiple documents that follow a valid reasoning trajectory. Moreover, their method requires intensive pretraining on the obtained corpus. \newcite{louvre2021} focus on weakly supervised multi-hop QA retrieval, yet their method uses corpus-specific (e.g., Wikipedia) heuristics and also requires pretraining. 
This motivates the need for data-efficient multi-hop retrieval methods that \textbf{(i)} work out-of-the-box without requiring additional (pre)training, and \textbf{(ii)} do not rely on hand-designed heuristics for data collection and annotation.

To this end, we present \textbf{\model}, which leverages the power of large language models (LLMs) for few-shot multi-hop retrieval. \model combines a simple unsupervised retrieval method i.e., TF-IDF similarity, with an LLM reranker that scores the relevance of document paths to a question based on the \textit{conditional likelihood of generating the question given the path}. 
Our approach makes use of instruction-based prompting \cite{t02021,ouyang2022training} to steer the LLM towards assigning higher scores to more relevant support document chains.\footnote{We use \textit{path} and \textit{chain} interchangeably throughout the paper. 
} 
To calibrate the model's reranking scores and alleviate prompt sensitivity \citep{calibrate2021}, we borrow techniques from the literature such as \textit{temperature scaling} \cite{kull2019beyond} and \textit{instruction ensembling} \citep{pet2021}. 
We also utilize \textit{demonstration ensembling} to leverage more examples than what can fit into the context of transformer LLMs by combining reranking probabilities computed with different demonstrations.

We evaluate few-shot \model on HotpotQA~\citep{hotpotqa}, a standard MQA benchmark, and show that it compares favorably against state-of-the-art models while using orders of magnitude fewer examples. More precisely, with only 128 training examples, \model outperforms DrKit \citep{drkit2020} and is only 4.2 Recall@10 points lower than multi-hop dense retrieval (MDR) \citep{mdr2021} (see \Cref{fig:teaser}).
We also showcase \model as part of a QA pipeline, again, displaying close QA performance to fully-supervised retrievers---only 4.1 F$_1$ points lower than MDR.

In summary, our contributions in this paper are:

\begin{enumerate}[leftmargin=*,noitemsep,topsep=0pt]
\item We propose \model, a few-shot reranking approach for multi-hop QA that reranks a given document path based on the likelihood of generating the question given a path prompt.

\item \model exhibits strong few-shot retrieval performance with as few as 128 examples and compares favorably to fully supervised methods (\Cref{sec:retrieval-eval}).

\item \model leads to strong QA performance when combined with a pretrained reader module, performing close to fully-supervised retrievers (\Cref{sec:qa-eval}).

\end{enumerate}

\cutsectionup
\section{\model}
\cutsectiondown

An overview of the full retrieval system is displayed in \Cref{fig:overview-prompt}: 
Given a question $q$, the system expands sequences of supporting documents into paths of length $H$, which are used to answer the question. 
At each step, we first use TF-IDF similarity to obtain an initial set of supporting document paths.\footnote{\model is agnostic to the retrieval approach and can be combined with other retrieval techniques. 
} 
We then use \model to rerank the current document chains based on their relevance to the question (\Cref{sec:ranking}).

%
Concretely, we start with retrieving $F$ candidate documents using TF-IDF for the `first hop'.
These `1-hop' candidates are scored by \model and $K_1$ top-ranked documents are kept and further expanded based on their hyperlinks to obtain 2-hop reasoning paths.\footnote{We assume the presence of hyperlinks following previous work \cite{Asai2019,qi2021answering} although \model is agnostic to how a candidate path is obtained.} 
These 2-hop reasoning chains are again reranked and the most promising $K_2$ candidates are further expanded. 
The process repeats until we obtain paths of length $H$, where $H$ can be a hyperparameter.\footnote{This process can be viewed as a variation of beam search.}
As the document graph can have a high branching factor, we only keep the top-$L$ hyperlinks as reranking candidates based on TF-IDF similarity between the hyperlink document and the question.
We have found this pruning step to improve efficiency without much performance drop. This process is shown in \Cref{fig:overview-prompt}(a).

\cutsubsectionup
\subsection{Path Reranking with \model}
\cutsubsectiondown
\label{sec:ranking}
Given a question $q$ and a reasoning path or chain $c$, we use an LM to score $c$ according to its relevance to $q$.
Concretely, we measure \textit{the likelihood of the question} given the path as follows:
\vspace{-0.5em}
\begin{equation}
\label{eq:score}
\text{Score}_q(c) = P_{\text{LM}}(q | \tau_c)
\end{equation}
where $P_{\text{LM}}(q | \tau_c)$ is the conditional probability of generating the question given a prompt $\tau_c$ containing path $\tau_c$ using an LM. 
%
Our initial experiments show that using $P_{\text{LM}}(q | \tau_c)$ works \textit{substantially} better than $P_{\text{LM}}(c | \tau_q)$ for a question-containing prompt $\tau_q$, which agrees with the findings in \citet{dos2020beyond}.\footnote{Earlier experiments showed that the recall of $P_{\text{LM}}(q | \tau_c)$ was at least 60\% better than that of $P_{\text{LM}}(c | \tau_q)$.} 
We argue that two factors contribute to this gap.
First, LMs can be sensitive to the \textit{surface form} \citep{holtzman-etal-2021-surface} of reasoning paths, making it difficult to reliably compare the probabilities of different reasoning paths using $P_{\text{LM}}(c | \tau_q)$. For instance, $P_{\text{LM}}(c | \tau_q)$ tends to be higher for shorter paths.
On the other hand, $P_{\text{LM}}(q | \tau_c)$ does not suffer from this issue since we compare the probabilities of the same string (i.e., the question) by conditioning on different reasoning paths. 
Second, the prompt format using $P_{\text{LM}}(q | \tau_c)$---the question follows a document---agrees more with the web data used for LM pretraining, where documents are usually followed by FAQs, questionnaires, and surveys, rather than the other way around.

We further add a \textit{temperature parameter} to scale the model output logits before computing $P(q | \tau_c)$. This can be seen as an instance of model calibration \cite{guo2017calibration,desai20,jiang2021can} with the goal of improving the reranking scores. We show that temperature scaling boosts reranking performance in \Cref{sec:retrieval-eval}. 
\cutparagraphup
\paragraph{Constructing Prompt $\tau_c$} 
As shown in \Cref{fig:overview-prompt} (b), the prompt consists of an instruction along with the document path. The instruction's goal is to encourage higher scores for more relevant paths by eliciting the LM reasoning ability \cite{ouyang2022training}. 
We note that the instruction part is \textit{fixed} across all prompts constructed for different candidate paths. 

The path is expressed in the prompt by \textit{concatenating} all documents in the chain and prepending each document with a fixed prefix, such as \textit{``Document:''} or \textit{``Passage:''}. 
The concatenation of path documents significantly improves reranking by \textit{simultaneously} considering all hops, which allows the LM to do a context-aware evaluation of path relevance. 

\cutsubsectionup
\subsection{Instruction Search and Ensembling}
\cutsubsectiondown
\label{sec:prompt-search}
\label{sec:ensembling-main}
Although instructions can be manually engineered to trigger the LM to accomplish the task (e.g., \emph{``Read the following documents and generate a question''}), this requires human expertise and can be sub-optimal. 
Therefore, we leverage automated instruction search \citet{gao2021}, where we use an encoder-decoder LM, e.g., a \tbase~model \cite{t5}, that is trained to fill masked text spans to generate instructions. 

Specifically, we fill in the template \textit{``Task: \texttt{<X>} documents and \texttt{<Y>} question. Question:''}, where \texttt{<X>} and \texttt{<Y>} are the masked spans expected to be filled in by the model (e.g., for a human-written instruction example, \texttt{<X>} = ``\emph{Read the following}'' and \texttt{<Y>} = ``\emph{answer the}'').
We consider two variations of this template corresponding to the cases where the document path appears before/after the template.
We constrained the template to contain the words `documents' and `question' to ensure that the model generates relevant prompts. We have found that using a less specific template without such tokens leads to more diverse but less relevant instructions. The exact templates used are in \Cref{app:prompt-search}.
 
Previous work has shown that mixing multiple prompts can improve few-shot performance \cite{gao2021,genpet2021}.
Similarly, such ensembling could produce more regularized path scores by alleviating prompt sensitivity \cite{calibrate2021}. Given a path, we combine the scores of obtained through different instructions.
We experiment with both \emph{mean} and \emph{max} ensembling.
\cutsubsectionup
\subsection{Demonstration Ensembling} 
\cutsubsectiondown
\label{sec:in-ctx}

We employ in-context learning (ICL) \citep{brown2020language} to teach the LLM to do reranking by showing the model examples i.e., demonstrations of questions and their gold paths.
A major obstacle to this approach is the \textit{input length limit} in standard transformer LMs. Since paths are comprised of multiple documents, in most cases we cannot feed more than two demonstrations without exceeding the limit of 1024 tokens, a standard setup for pretrained LMs. To workaround that, we utilize \textit{demonstration ensembling}, where different in-context demonstrations are used to compute scores for a given path, and the scores are combined by a mean or max operation. 

\cutsectionup
\section{Experiments}
\cutsectiondown
\paragraph{Data}
We evaluate our method on \textbf{HotpotQA} \cite{hotpotqa}, which consists of two-hop questions over diverse topics. We focus on the \textit{fullwiki} setting in which two Wikipedia passages are required to answer the questions. Since the gold passages for the test set are not available, we follow prior work and evaluate \model on the development set, which has 7,405 questions. 
There are two main question types in HotpotQA: 
(1) \textit{comparison} questions usually require contrasting two entities and (2) \textit{bridge} questions can be answered by following a connecting entity that links one document to another. We also evaluate few-shot \model on the \textbf{2WikiMQA} dataset \citep{ho-etal-2020-constructing} in \Cref{app:wikimqa}.

\paragraph{Compute Infrastructure}
All our reranking experiments are run on a workstation with a single Nvidia A40 GPU and 256GB of RAM. Our QA experiments in \Cref{sec:qa-eval} are run on a workstation with two Nvidia Quadro RTX 8000 GPUs and 128GB of RAM. 
\cutparagraphup
\paragraph{Models}
We use HuggingFace implementations \cite{wolf2020transformers} of \gptxl~(1.5B) \cite{brown2020language}, \tbase~(220M), \tlarge~(770M) and \txl~(3B) \cite{t5} in our experiments.
We use the `LM adapted' version of T5 models since they have been shown to work better for prompt-based learning \cite{lester21}. We report additional results with the OPT-30B model \citep{zhang2022opt} in \Cref{sec: opt}.

\cutparagraphup
\paragraph{Hyperparameters}
For \model, we use a path length of $H=2$ for all experiments. For pruning the search space we use $K_1=5$ and $L=3$. We use the TF-IDF index implemented by \citet{Asai2019} and initially retrieved $F=100$ documents from TF-IDF. 
We truncate path documents to 230 tokens before constructing the prompt and limit the prompt length to 600 tokens. When using in-context demos, we use the maximum length of 1024 tokens.


\cutparagraphup
\paragraph{Metrics}
Retrieval performance is measured using both Recall (R$\text{@}k$) and Answer Recall (AR$\text{@}k$), with $k \in \{2, 10, 20\}$. R$\text{@}k$ measures whether the two gold documents are present in the top-$k$ retrieved documents and AR$\text{@}k$ is the recall of the answer string in the top-$k$ retrieved documents.
For HotpotQA, we only compute AR over questions with span answers (we ignore yes/no and comparison questions). Since we do not have access to the HotpotQA test set, we report results on the original development set provided by \citet{hotpotqa}. 

\cutparagraphup
\paragraph{Document Scores} 
We compute document scores from path scores as follows.
Similar to \citet{das2019chains}, we take a document score to be the \textit{maximum} of all its path scores. We find this change to yield better recall than using path scores, with details elaborated in \Cref{app:sec:doc-scores}.

\cutparagraphup
\paragraph{Instruction Search and Temperature }
For instruction search, we generate 200 different instructions as described in \Cref{sec:prompt-search} using top-$k$ sampling with $k=10$. Then, we select the best instruction based on R@2 over our development set of 128 examples. The same process is used to select the optimal temperature value. 
\Cref{app:tab:prompts-app} shows the best 10 instructions identified.

\begin{table*}[ht!] 
\centering
\footnotesize
\begin{tabular}{lccccccc}
\toprule
 & \textbf{\# Ex.} \phantom{\tiny(.0)}&    \textbf{R@2} \phantom{\tiny(.0)} &  \textbf{R@10} \phantom{\tiny(.0)}&  \textbf{R@20} \phantom{\tiny(.0)}&  \textbf{AR@2} \phantom{\tiny(.0)}&  \textbf{AR@10}\phantom{\tiny(.0)} &  \textbf{AR@20}\phantom{\tiny(.0)} \\
  \midrule

 \textbf{\textit{Unsupervised Baselines}} &&&&&&\\
 TF-IDF         & --\phantom{\tiny(.0)} &  9.9 \phantom{\tiny(.0)} & 27.6 \phantom{\tiny(.0)}  & 35.0 \phantom{\tiny(.0)}  & 37.6 \phantom{\tiny(.0)} & 53.8 \phantom{\tiny(.0)}   & 60.2 \phantom{\tiny(.0)} \\
 TF-IDF + BM25  & --\phantom{\tiny(.0)} & 19.1  \phantom{\tiny(.0)} & 54.7 \phantom{\tiny(.0)}  & 61.8 \phantom{\tiny(.0)}  & 49.5 \phantom{\tiny(.0)} & 74.7 \phantom{\tiny(.0)}   & 79.9 \phantom{\tiny(.0)} \\
 \midrule
 
 \textbf{\textit{Fully-supervised Baselines}} &&&&&&\\
 DrKit & \textasciitilde 90K           \phantom{\tiny(.0)} & 38.3 \phantom{\tiny(.0)} & 67.2 \phantom{\tiny(.0)} & 71.0 \phantom{\tiny(.0)} & --  \phantom{\tiny(.0)}  & --  \phantom{\tiny(.0)}   & --   \phantom{\tiny(.0)}  \\ 
 MDR & \textasciitilde 90K             \phantom{\tiny(.0)} & 65.9 \phantom{\tiny(.0)} & 77.5 \phantom{\tiny(.0)} & 80.2 \phantom{\tiny(.0)} & --  \phantom{\tiny(.0)}  & --  \phantom{\tiny(.0)}   & --   \phantom{\tiny(.0)}  \\
 PathRetriever & \textasciitilde 90K   \phantom{\tiny(.0)} & 66.4 \phantom{\tiny(.0)} & 77.8 \phantom{\tiny(.0)} & 78.7 \phantom{\tiny(.0)} & 82.2\phantom{\tiny(.0)}  & 90.5\phantom{\tiny(.0)}  & 90.5  \phantom{\tiny(.0)}  \\
 
\midrule
\textbf{{\model}, \textit{no ICL}} &&&&&&\\


\gptxl\textsuperscript{\textdagger} & --   \phantom{\tiny(.0)} & 36.6 \phantom{\tiny(.0)} &  60.5     \phantom{\tiny(.0)}  &  65.9  \phantom{\tiny(.0)}    &  63.0    \phantom{\tiny(.0)}  &   83.9   \phantom{\tiny(.0)}  &   87.4   \phantom{\tiny(.0)}  \\ 
\txl\textsuperscript{\textdagger}   & --   \phantom{\tiny(.0)} & 42.8 \phantom{\tiny(.0)} &   68.9    \phantom{\tiny(.0)}  &  74.1  \phantom{\tiny(.0)}    &   69.3   \phantom{\tiny(.0)}  &   86.8   \phantom{\tiny(.0)}  &  89.0    \phantom{\tiny(.0)}  \\ 
\hspace{0.1cm} + best inst.         & 128  \phantom{\tiny(.0)} & 47.8 \phantom{\tiny(.0)} &   71.4    \phantom{\tiny(.0)}  &  76.0  \phantom{\tiny(.0)}    &   74.0   \phantom{\tiny(.0)}  &   87.9   \phantom{\tiny(.0)}  &  89.7    \phantom{\tiny(.0)}  \\
\hspace{0.1cm} + temp. scaling      & 128  \phantom{\tiny(.0)} & 49.7 \phantom{\tiny(.0)} &   71.9    \phantom{\tiny(.0)}  &  76.2  \phantom{\tiny(.0)}    &   76.2   \phantom{\tiny(.0)}  &   88.4   \phantom{\tiny(.0)}  &  89.9    \phantom{\tiny(.0)}  \\
\hspace{0.1cm} + inst. ensemble     & 128  \phantom{\tiny(.0)} & 51.3 \phantom{\tiny(.0)} &  {72.0}   \phantom{\tiny(.0)}  &  76.4  \phantom{\tiny(.0)}    &   {77.6} \phantom{\tiny(.0)}  &   {88.5} \phantom{\tiny(.0)}  &  {90.3}  \phantom{\tiny(.0)}  \\


\midrule 
\textbf{{\model}, \textit{with ICL}} &&&&&&\\




\txl, $N_{\text{demos}}=2$  & 128  \phantom{\tiny(.0)} &  52.3 \tiny(.7)  &  73.1 \tiny(.2) &  \textbf{77.1} \tiny(.2) &  78.6 \tiny (.7) &   88.7 \tiny (.0)  &   90.3 \tiny (.1) \\
\txl, $N_{\text{demos}}=8$  & 128  \phantom{\tiny(.0)} & \textbf{54.5} \tiny(.7)  & \textbf{73.6} \tiny(.3) & 76.9 \tiny(.1) & \textbf{79.1} \tiny(.6) & \textbf{89.0 } \tiny(.1)  & \textbf{90.5} \tiny(.0) \\
\txl, $N_{\text{demos}}=10$  & 128  \phantom{\tiny(.0)} & 54.4 \tiny(.5)  & {73.5} \tiny(.3) & 76.9 \tiny(.1) & 78.9 \tiny(.4) & 88.9 \tiny(.1)  & \textbf{90.5} \tiny(.0) \\




\bottomrule
\end{tabular}
\cutcaptionup
\caption{
\footnotesize
Retrieval performance on HotpotQA comparing \model to baselines. \textdagger: No instruction used. \model results except those marked with \textdagger{} use a labeled set of 128 examples for tuning the instruction and the temperature parameter. Few-shot experiments use the best instruction found on a held-out set of 128 examples (See \Cref{app:tab:prompts-app} in Appendix) and temperature ($T=1.4$).  In-context learning (ICL) experiments are run 5 times with demos sampled from the same 128-example set and we report mean and (std). Our best results are highlighted in \textbf{bold}.
}
\cutcaptiondown
\label{tab:retrieval-hotpot}
\end{table*}
\cutparagraphup
\paragraph{Baselines} 
We compare our reranker to the following baselines.  
\textbf{TF-IDF} retrieves top similar documents to the question using TF-IDF similarity and \textbf{TF-IDF + BM25} adds an extra step where retrieved documents and their hyperlinks are reranked using BM25 \cite{robertson1995okapi}.
\textbf{PathRetriever} \cite{Asai2019} is a graph-based retriever trained to expand an initial pool of documents based on Wikipedia links
and searches for the best path using beam search.\footnote{We run PathRetriever on HotpotQA with original hyperparameters except for an initial TF-IDF pool size=100 to allow for fair comparison to our approach.}
\textbf{DrKIT} \cite{drkit2020} is an end-to-end trained dense retrieval approach that starts from question entities and traverses a virtual knowledge base to find the relevant entities.
Multi-hop Dense Retrieval \textbf{(MDR)} \cite{mdr2021} encodes the question and the documents retrieved by each step into a dense vector and uses maximum inner-product search (MIPS) to find the next hop. 

Below, we start with the evaluation of the zero- and few-shot reranking of \model (\Cref{sec:retrieval-eval}).
Then, we move to evaluate downstream MQA performance in the few-shot setting (\Cref{sec:qa-eval}).

\cutsubsectionup
\subsection{Retrieval Performance}
\cutsubsectiondown
\label{sec:retrieval-eval}
\Cref{tab:retrieval-hotpot} shows the performance of \model and other comparisons in zero- and few-shot settings. 

\cutparagraphup
\paragraph{Zero-shot Performance} 
We start with discussing the retrieval performance of zero-shot \model on HotpotQA. 
First, we observe that simple TF-IDF performs poorly in terms of different recall metrics, while TF-IDF + BM25 performs much better, yet still worse than fully-supervised approaches. 
Next, we look at the performance of the zero-shot \model (\txl) which uses no instructions, i.e., the prompt consists of only the document path. 
These models obtain better recalls than TF-IDF + BM25 and even outperform the fully-supervised DrKit. 
Although this approach does not use \textit{any} labeled data, it is only 3.7 AR@10 points worse than PathRetriever, which is trained on \textasciitilde$90$K examples. These findings demonstrate \model's effectiveness at reranking paths of documents. 

\begin{figure}[t!]

    \centering
    \hspace*{-5pt}
    \includegraphics[width=6.8cm]{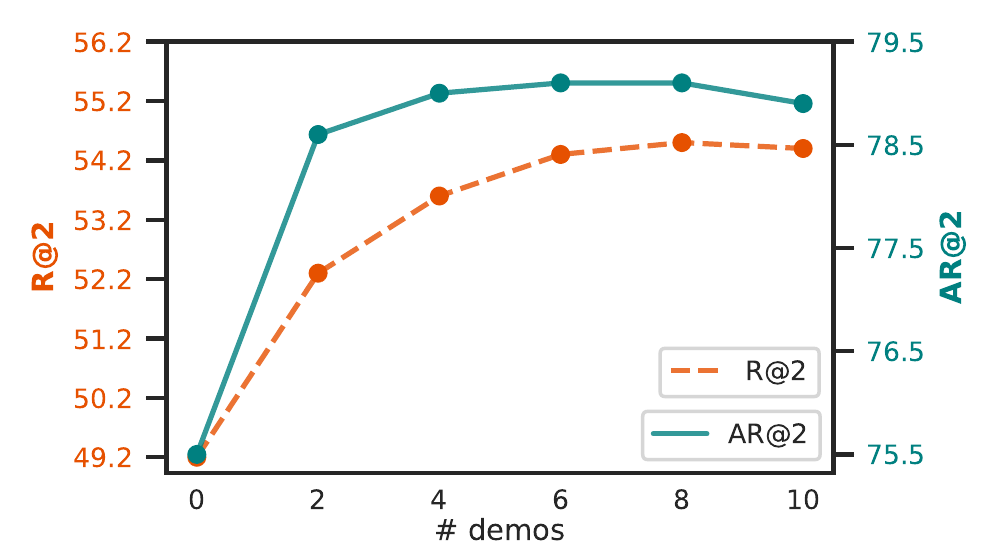}
    \cutcaptionup
    \caption{
    \footnotesize
    Demonstration ensembling (\Cref{sec:in-ctx}) is able to leverage more examples ($N > 2$) than what is allowed by the ~\txl~ context window size. We show R@2 and AR@2 on HotpotQA with different numbers of demonstrations.  Metrics are averaged over 5 runs with different demos sampled from a 128-example set.
    }
    \cutcaptiondown
    \label{fig:fewshot}
\end{figure}

\cutparagraphup
\paragraph{Few-shot Performance} 
The zero-shot performance of \model can be further improved with access to a small set of labeled examples (in our case, we only used 128 examples from HotpotQA) for instruction search and finding temperature value. We observe a substantial boost of 11.6\% (42.8 $\rightarrow$ 47.8) in R@2 of \model when using the \textit{best instruction} found by instruction search. Furthermore, temperature scaling with $T=1.4$ also provides a boost of 3.9\% (47.8 $\rightarrow$ 49.7) points in R@2. 
We also observe that instruction ensembling gives a further performance boost, reaching 51.3 R@2 with \model. We show the performance of \textit{max ensembling},
which we have found to perform better than mean ensembling in terms of R@2. We hypothesize that max ensembling computes an \textit{upper bound} on the path scores, compensating for any underestimation of path scores that can happen when using a single instruction.

\cutparagraphup
\paragraph{In-context learning} We experiment with an $N$-shot setting while making sure that the two demonstrations cover \textit{both} question types in HotpotQA (bridge and comparison). \Cref{fig:fewshot} shows that both R@2 and AR@2 improve as we use more demonstrations. With only 2 examples, we observe a large boost of 6.3\% (49.2 $\rightarrow$ 52.3) in R@2. Since we cannot fit more than 2 demonstrations in the 1024 context window, we use demonstration ensembling (\Cref{sec:in-ctx}). 
For instance, 6-shot ensembling scores a path by combining 3 different contexts, each obtained using 2 demonstrations.
We use max ensembling as it is found to work best.
\Cref{fig:fewshot} shows the in-context learning performance with a different number of demonstrations.
We observe a steady increase in R@2 until $N=8$. AR@2 also improves with more demonstrations but drops slightly with $N=10$.
Interestingly, demonstration ensembling has enabled us to leverage more examples than permitted by the context window size of T5-XL. We leave it to future work to study the applicability of this technique to other tasks.

\cutsubsectionup
\subsection{Full QA Performance}
\cutsubsectiondown
\label{sec:qa-eval}
We analyze the performance of \model when used as the retriever in a QA pipeline. We adopt an extractive reader model based on ELECTRA Large \cite{electra2020} with two heads to predict the start and end of the answer span. We use the checkpoint provided by \citet{mdr2021}, and the same inference setting.
Details on the inference hyperparameters for the reader are in \Cref{app:sec:electra}.

In \Cref{tab:qa-eval}, we compare the QA performance on HotpotQA \textbf{development set} with \model as the retriever against a \textit{fully-supervised} retriever, namely MDR \citep{mdr2021} as well as unsupervised TF-IDF. 
\model with $N_{\text{demos}}=10$ is only 4.6 F$_1$ points worse than MDR, which is using the \textit{same} reader module. \Cref{tab:qa-eval-test} shows performance on HotpotQA \textbf{test set} with different \textit{fully-supervised} systems compared to \model ($N_{\text{demos}} = 2)$, where \model is only 1.9 and 4.2 EM points below PathRetriever and MDR, respectively.


\begin{table}[]
\footnotesize
    \centering
    \begin{tabular}{lll}
    \toprule
    \textbf{Retriever}     &   \textbf{EM} & \textbf{F1} \\
       \midrule
  \textbf{\textit{Fully-supervised}} \\
  MDR \citep{mdr2021} & 62.3 \phantom{\tiny(.2)} & 75.1 \phantom{\tiny(.2)} \\
  \midrule
  \textbf{\textit{Zero-shot}} \\ 
 TF-IDF& 39.6\phantom{\tiny(.0)} & 49.4 \phantom{\tiny(.0)} \\
 \model, no inst& 55.7 \phantom{\tiny(.0)} & 67.7 \phantom{\tiny(.2)} \\
 \midrule
   \textbf{\textit{Few-shot}} \\ 
  \model, ($N_{\text{demos}}=2$) & 57.8 \tiny(.1) & 70.0 \tiny(.1) \\
  \model, ($N_{\text{demos}}=10$) & 58.3 \tiny(.0) & 70.5 \tiny(.1) \\

\bottomrule
    \end{tabular}
    \cutcaptionup
    \caption{
    \footnotesize
    Answer EM and F1 on HotpotQA development set. 
    \model results are aggregated over 3 runs with different demonstrations. We show metrics mean and (std). 
    To allow for a fair comparison, only the retriever is varied over these systems while the reader module is the \textit{same}.
    }
    \cutcaptiondown
    \label{tab:qa-eval}
\end{table}

\begin{table}[]
\footnotesize
    \centering
    \begin{tabular}{lll}
    \toprule
    \textbf{Retriever}     &   \textbf{EM} & \textbf{F1} \\
       \midrule
  DrKit \citep{drkit2020} & 42.1 & 51.7\\
  PathRetriever \citep{Asai2019} & 60.0 & 73.0 \\
  MDR \citep{mdr2021} & 62.3 & 75.3 \\
  \midrule 
  \model, ($N_{\text{demos}}=2$) & 58.1 & 71.1  \\

\bottomrule
    \end{tabular}
    \cutcaptionup
    \caption{
    \footnotesize
    Answer EM and F1 on HotpotQA test set. 
    MDR and \model use the same ELECTRA reader, while other systems use different readers. 
    }
    \cutcaptiondown
    \label{tab:qa-eval-test}
\end{table}

\section{Analysis}
\label{sec: analysis}

\cutsubsectionup
\subsection{Comparison to Single-hop Reranking}
\cutsubsectiondown
\label{sec:single-hop}        
The key idea behind our approach is to conduct joint reasoning with documents in the path using the LM, as opposed to reranking each document in the path separately (\textit{single-hop reranking}).
More specifically, in single-hop reranking, we expand paths using the same setup of \model but rerank each document $d$ separately using $p(q | \tau_d)$, for a given document prompt $\tau_d$. 

To assess whether our multi-hop reranking approach offers the advantage of global reasoning, we compare both approaches by running two experiments with identical settings except for how documents are reranked. 
For evaluation, we use a set of 4K questions from HotpotQA and \tlarge, and no instruction is used, i.e., the prompt only contains the document(s). \Cref{tab:single-hop} shows the retrieval performance of both approaches. Interestingly, a large gap in recall scores is observed between single-hop and multi-hop reranking. This supports our hypothesis that jointly considering multiple documents in the path helps the LM better model documents' relevance to the question.

\begin{table}[]
\footnotesize
    \centering
    \vspace*{0.06in}
    \begin{tabular}{lcccc}
    \toprule
    \textbf{Re-ranking}     &   \textbf{R@2} & \textbf{R@10} & \textbf{AR@2} & \textbf{AR@10}\\
       \midrule
    Single-hop &  22.8 &  52.0 & 54.9 & 73.8 \\ 
Multi-hop & 46.9 &  67.6 & 75.4 &   87.9 \\
\bottomrule
    \end{tabular}
    \cutcaptionup
    \caption{
    \footnotesize
    Recall measured on 4K questions from HotpotQA in two settings: reranking each document separately with the LM (single-hop) and reranking the full path at once (multi-hop). Multi-hop reranking performs substantially better than single-hop.
    }
    \cutcaptiondown
    \label{tab:single-hop}
\end{table}

\cutsubsectionup
\subsection{Role of Instruction}
\begin{figure}[t!]
    \centering
    \hspace*{-0.25cm}
    \includegraphics[width=0.465\textwidth]{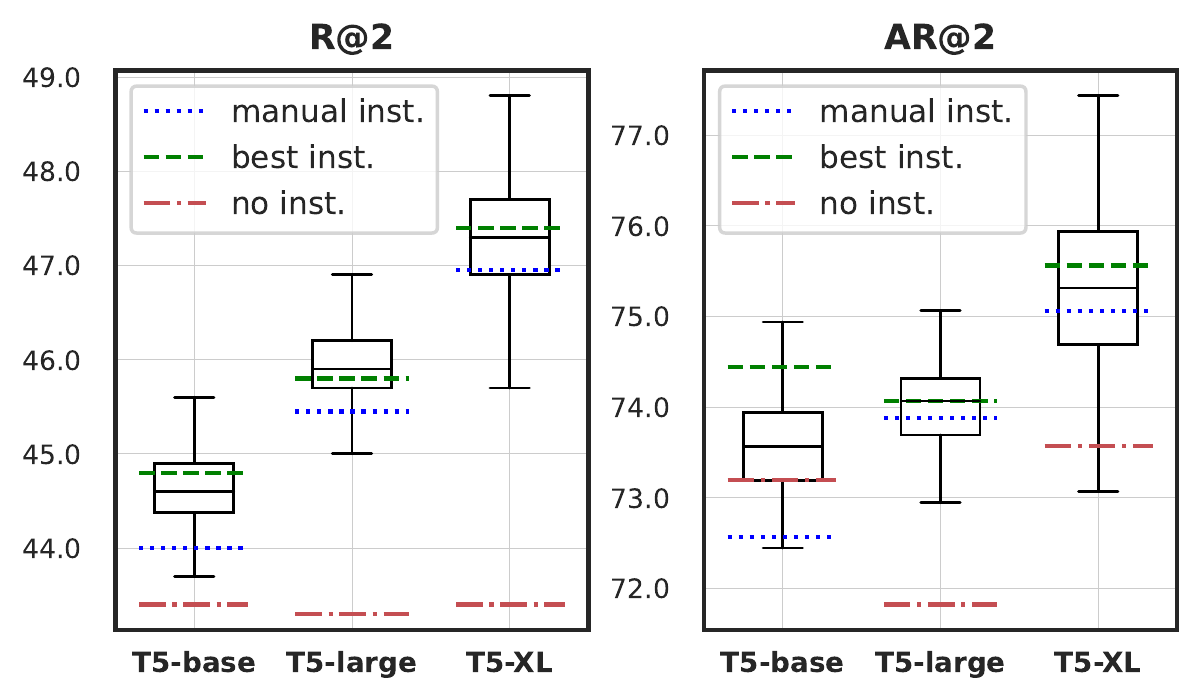}
    \cutcaptionup
    \caption{
    \footnotesize
    R@2 and AR@2 with different kinds of instructions for three different T5 sizes: XL, Large, and Base. The recall is measured over 1K questions from HotpotQA train set using 200 different instructions obtained using automated search \Cref{sec:prompt-search}.
    }
    \label{fig:prompt-sens}
\end{figure}

\cutsubsectiondown
\label{sec:sensitivity}
Our goal here is to investigate \textbf{(i)} how useful is the presence of the instruction in the prompt, \textbf{(ii)} how much benefit (if any) automated instruction search provides over manual instructions, and \textbf{(iii)} whether the instruction's location in the prompt matters. To answer these questions, we analyze the recall over 200 different instructions generated using the method described in \Cref{sec:prompt-search} and using 1K examples from HotpotQA with different LM sizes: \txl, \tlarge, and \tbase, with results displayed in \Cref{fig:prompt-sens}. This analysis uses an initial set of TFIDF documents of size $F=30$.

\cutparagraphup
\paragraph{Usefulness of Instruction}
We can see that using no instruction consistently yields poorer performance than using an instruction of any sort, across all variants of T5. Interestingly, without the instruction, the three model sizes have almost the same R@2. The difference in their performances becomes apparent when an instruction is added. 
Strikingly, in the no instruction case, \tlarge~performs \textit{worse} than \tbase~in terms of AR@2, showing that scaling does not consistently help recall when no instructions are used. This hints at the fact that instructions play a major role in harnessing the full power of LLMs, at least for our task. 


\cutparagraphup
\paragraph{Benefit of Automated Instruction Search}

Next, we compare a human-written instruction against an instruction found through automated instruction search on a labeled set of 128 examples.
The manual instruction we use is \emph{``Please write a question based on these passages.''}, which is used by \citet{sachan2022improving}.\footnote{We average recall of the two cases where the instruction falls before and after the path. See the next paragraph for more context.} In \Cref{fig:prompt-sens}, we compare the recall when using these instructions. Interestingly, the search-based instruction outperforms the manual one in almost all cases. We also observe that the manual instruction performs poorly for AR@2 on T5-base, even worse than no instruction. These observations hint at the utility of automated instruction search for path reranking. However, it is worth noting that the best instruction on a relatively small held-out set will not necessarily generalize during test time: The search-based instruction produces AR@2 and R@2 that are almost the same or worse than the median instruction, respectively with T5-Large.

\begin{figure}[t!]
    \centering
    \hspace*{-1pt}
    \includegraphics[width=0.47\textwidth]{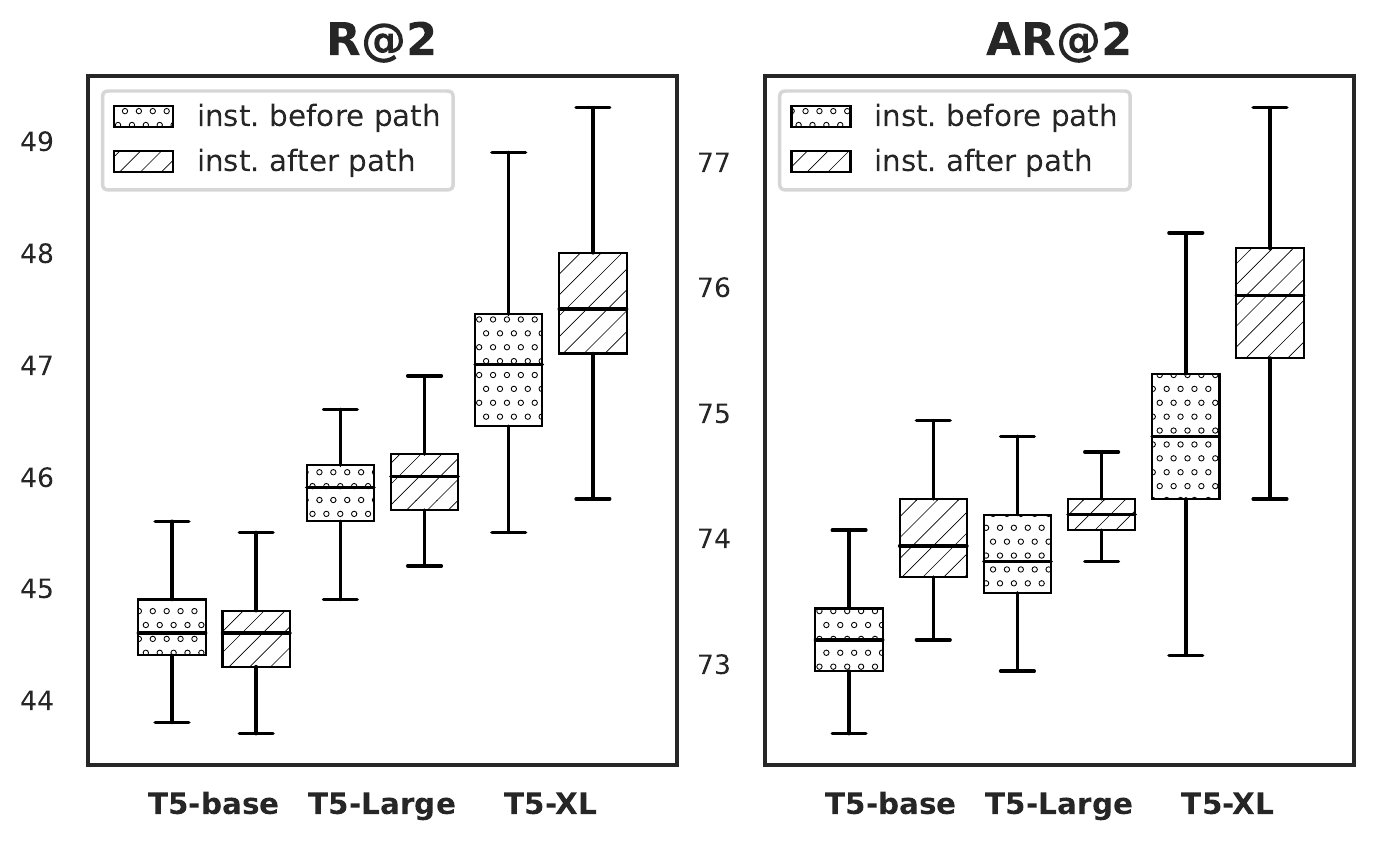}
    \cutcaptionup
    \caption{
    \footnotesize
    Retrieval performance when placing the instruction \textit{before} and \textit{after} the document path in the prompt. The recall is measured over 1K questions from HotpotQA train set using 200 different instructions. Having the instruction after the path performs consistently better which is likely due to recency bias \cite{calibrate2021}. 
    }
    \cutcaptiondown
    \label{fig:inst-order-sens}
\end{figure}

\cutparagraphup
\paragraph{Location of Instruction}
We study the performance of two different kinds of prompts, where the instruction appears \textit{before} and \textit{after} the path. \Cref{fig:inst-order-sens} shows the R@2 and AR@2 in both cases for T5 models of different sizes. 
We observe that placing the instruction after the path performs \textit{consistently better} than placing before it, across all model variants. We hypothesize this to be an instance of the \textit{recency bias} exhibited by LMs \cite{calibrate2021}, \textit{i.e.,} placing the instruction right before where the model is asked to generate the question better primes the LM for the task and produces better calibrated path scores. We expect such finding to generalize to other tasks where instruction-based prompting is used. 



\subsection{Choice of Language Model}
\cutsubsectiondown
\label{sec: opt}

\begin{table}[t!]
\footnotesize
    \centering
    \begin{tabular}{ccccc}
    \toprule
    \textbf{Model}     &   \textbf{R@2} & \textbf{R@10} & \textbf{AR@2} & \textbf{AR@10}\\
    \midrule
    OPT-30B & 36.9 & 65.4 & \textbf{61.0} & 82.0 \\
    GPT2-XL & \textbf{47.2} & \textbf{70.3} & 57.1 & \textbf{85.7} \\
    \bottomrule
    \end{tabular}
    \cutcaptionup
    \caption{\footnotesize
    Document and answer recall of GPT2 and OPT models based on 1000 questions from HotpotQA.}
    \cutcaptiondown
    \label{tab:optexp}
\end{table}

\Cref{tab:optexp} compares the reranking performance of GPT2-XL and OPT-30B \citep{zhang2022opt} models.
Despite having an order of magnitude more parameters, we observe that the OPT model is generally worse compared to the smaller GPT2-XL model.
We suspect this is due to domain mismatch between pre-training data and task relevant data.
Pre-training data of GPT2 models is potentially more biased towards Wikipedia data compared to the OPT models which are trained on more diverse data.
Importantly, this shows that scaling up the language model doesn't necessarily guarantee better reranking performance and domain gap is an important consideration.

\cutsubsectionup
\subsection{Further Analysis and Comparisons}
\label{sec:inf-cost}
\cutsubsectiondown
We further analyze the inference cost of \model compared to PathRetriever and MDR in \Cref{app:inf-cost}. In \Cref{app:doc-order}, we study \model's recall sensitivity to document order in the prompt $\tau_c$ by comparing performance using two different document ordering schemes in the prompt. Lastly, we compare \model to few-shot PathRetriever and LOUVRE \citep{louvre2021} in \Cref{app:few-shot-compare}.

\cutsectionup
\section{Related Work}
\cutsectiondown

\paragraph{Multi-hop Retrieval} 
The majority of approaches for multi-hop question answering rely on two main components: a retriever and a reader. The retriever component can be a sparse index or heuristic-based such as TF-IDF or BM25 \cite{chen17,semret20} or dense index \cite{dpr2020,mdr2021,beamdr21}. Other approaches aimed to improve the retriever with an additional \textit{reranking} step on top of a simple retriever \cite{wang2018r,lee18,htut18}. \citet{Asai2019} combined TF-IDF retriever with a recurrent graph retriever and used the reader module to rerank paths based on answer confidence. \citet{qi2021answering} used a single transformer model to perform retrieval, reranking and reading in an iterative fashion. However, the good performance of previous work comes mainly from training on a large number of examples and is likely to fail in low-data settings. To treat this issue, \citet{louvre2021} proposed to pretrain MDR \cite{mdr2021} on a large number of weakly-supervised examples of questions and the corresponding document paths. Although promising in low-data settings, their pretraining is computationally expensive as it is done on millions of examples. On the other hand, our approach requires no task-specific pretraining.

\cutparagraphup
\paragraph{Language Models Prompting } Prompt-based learning aims to construct better inputs, \textit{i.e.,} prompts to language models to elicit better zero- or few-shot performance \cite{brown2020language,liu2021pre}. Recently, instruction tuning, where a language model is trained to follow natural language instruction, has shown impressive zero-shot performance on unseen tasks \cite{flan2021,ouyang2022training}. In our work, we use instructions to guide to model toward assigning better scores to more relevant document paths. 
\cutparagraphup
\paragraph{LM-based Reranking}
Our scoring function is related to query likelihood retrieval \citep{lavrenko2017relevance,  ponte2017language} and is in line with previous work that employed generative language models for passage reranking \citep{nogueira20}. 
\citet{dos2020beyond} performed single-hop reranking using question likelihood given the passage, but their setting was limited to fully-supervised, single-hop QA. 
Concurrent with our work is \citep{sachan2022improving}, where the authors leverage LLMs for unsupervised passage reranking for QA. While their focus is on single passages, we study the reranking of multi-passage paths, which is more challenging. Moreover, their exploration of prompting is limited to a single manual instruction, whereas we provide an in-depth analysis of the effect of different prompting aspects on the recall such as instruction importance, location in the prompt, and manual vs. automated. 

\cutsectionup
\section*{Conclusion}
\cutsectiondown
We introduced \model, a method to perform few-shot reranking of multi-document paths for multi-hop question answering based on large language models. 
Experiments on a standard multi-hop QA benchmark show the strong performance of \model in the few-shot setting compared to fully-supervised multi-hop reranking systems. 
Future avenues of exploration include combining \model with efficient tuning techniques such as prefix tuning and efficient strategies for instruction search.


\cutsectionup
\section*{Limitations}
\cutsectiondown
One limitation to LM-based reranking is the computational overhead involved in reranking paths. Our approach requires a forward pass through the LM to rerank each path, which can become expensive when using relatively large models such as GPT-3 or when dealing with more hop count that creates combinatorially more paths. Another limitation of \model is imposed by the transformer context window length. Since \model requires the prompt to include all path documents, it could be infeasible to fit all path documents into the prompt for paths with a larger hop count. A potential direction to workaround this is to condense or summarize the path documents beforehand. We leave it to future work to explore this and other techniques.

\cutsubsectionup
\section*{Acknowledgements}
\cutsubsectiondown
This work is supported by LG AI Research. Additionally, we would like to thank Sewon Min for providing feedback on the paper. We also thank the anonymous reviewers for their valuable suggestions.

\bibliographystyle{acl_natbib}
\bibliography{ref}

\appendix
\clearpage

\setcounter{table}{0}
\renewcommand{\thetable}{A\arabic{table}}

\section{Instructions}
\label{app:sec:inst}

\subsection{Best Instructions}
\Cref{app:tab:prompts-app} shows the top 10 performing instructions found by instruction search (\Cref{sec:prompt-search}) based on R@2 and using T5-XL. 


\subsection{Instruction Search}
\label{app:prompt-search}
The actual templates we feed T5 are \textit{``Task: \texttt{<X>} documents \texttt{<Y>}} question based on them. Question:'' and \textit{``Task: \texttt{<X>} previous documents and \texttt{<Y>} question based on them. Question:''}. We have found using the phrase ``based on them'' to be essential in directing the model to generate sensible instructions. Otherwise, the model would generate something like ``Read the documents in question..''. However, we remove that phrase from the obtained instructions''.


\section{Document Scores}
\label{app:sec:doc-scores}

It is not immediately obvious how to compute a final score for each document since \model is mainly used to score document. The main issue is that a document can fall on multiple paths at the same time (some of which could be incomplete or not fully expanded yet) and therefore could have multiple such scores.

For example, assume a path $A \rightarrow B \rightarrow C$ of consisting of the documents $A$, $B$, and $C$, respectively. Considering the document $B$, we see that two scores are associated with $B$: score of the sub-path $A \rightarrow B$ and score of the full $A \rightarrow B \rightarrow C$ path. To compute the final score of $B$, we could either just take the score of the longest path, or combine the two scores using mean, minimum, or maximum operations. What we found to work best compared to other alternatives is to take \textit{maximum}, which is similar to what is done in \cite{das2019chains}. We use this formulation when computing our recall metrics in \Cref{sec:retrieval-eval}.

\section{Hyperparameters}

\subsection{ELECTRA Reader}
\label{app:sec:electra}
We use the same reader setting as in \citet{mdr2021}, where the top-100 retrieved paths are fed to the reader to obtain an answer from each path. Answers are then sorted based on a linear combination of path score and answer confidence, and the top answer is returned. We use the default hyperparameters for HotpotQA from \cite{mdr2021} in their codebase.\footnote{\url{https://github.com/facebookresearch/multihop\_dense\_retrieval}} We use a maximum path length of 512 tokens, maximum question length of 64, and answer length of 30. In their experiments, \citet{mdr2021} combine the answer confidence along with a ranking score using linear interpolation with a hyperparameter $\lambda$. For our experiments, we use the path scores produced by \model instead and learn $\lambda$ on a held-out development set. The value we end up using for $\lambda$ is 0.9.


\begin{table*}[h!]
\small
    \centering
    \begin{tabular}{c l}
    \toprule
        \textbf{ID} & \textbf{Prompt} \\ 
\midrule
1 \afterprompt{Review previous documents and ask some question.} 
\midrule
2 \afterprompt{Review the previous documents and answer question.} 
\midrule
3 \afterprompt{Read the previous documents and write the following question.} 
\midrule
4 \afterprompt{Search previous documents and ask the question.}
\midrule
5 \beforeprompt{To analyze the documents and ask question.}
\midrule
6 \afterprompt{To read the previous documents and write a question.}
\midrule
7 \afterprompt{Read previous documents and write your exam question.}
\midrule
8 \afterprompt{Read the previous documents and ask this question.}
\midrule
9 \beforeprompt{Read two documents and answer a question.}
\midrule
10 \beforeprompt{Identify all documents and ask question.}
\bottomrule
    \end{tabular}
    \caption{Top 10 instructions found  through automated instruction search (\Cref{sec:prompt-search}) using T5-XL. Instructions are sorted in descending order according to R@2 on a held-out development set of size 128 from HotpotQA \cite{hotpotqa}. We use the first 5 for instruction ensembling (section ~\Cref{sec:ensembling-main}). \textcolor{blue}{Blue} represents fixed text that does not depend on the path \textit{i.e} the instruction. The tokens \texttt{[D1]}, \texttt{[D2]},..,  etc. indicate where path documents are inserted. }
    \label{app:tab:prompts-app}
\end{table*}


\section{Further Results and Analysis}

\subsection{Results on 2WikiMQA}
\label{app:wikimqa}

\Cref{tab:wiki-mqa} shows the performance with few-shot \model with only one setting: few-shot, best instruction and with temperature scaling on the 2WikiMQA dataset \cite{ho-etal-2020-constructing}. We compare it to two unsupervised baselines, namely TF-IDF and TF-IDF+BM25 . \model is significantly outperforming the baselines while using only 128 examples for tuning the instruction and the temperature parameter.

  \begin{table*}[h!]
  \footnotesize
      \centering
      \begin{tabular}{lcccccc}
      \toprule
      
          &   \textbf{R@2} &  \textbf{R@10} &   \textbf{R@20} &   \textbf{AR@2} &   \textbf{AR@10} &   \textbf{AR@20} \\
         \midrule
          \textbf{\textit{Unsupervised Baselines}} &&&&&&\\
      TF-IDF & 5.8 &	8.9 &	23.8 &	22.2 &	37.8 &	44.2  \\
      TF-IDF+BM25 & 7.6	& 34.1 & 45.8	& 20.7 &	44.9 &	53.0  \\
      \midrule
      \textbf{{\model}, \textit{no ICL}} &&&&&&\\
      T5-XL, best inst., temp. scaling & 19.3 &	58.6 &	62.7 &	33.9 &	60.0 &	64.2 \\
  \bottomrule
      \end{tabular}
      \caption{
      \footnotesize
      Retrieval performance on 2WikiMQA \cite{ho-etal-2020-constructing}.
      }
      \label{tab:wiki-mqa}
  \end{table*}

\subsection{Inference Cost}
\label{app:inf-cost}
Here, we analyze inference cost in terms of latency per query. We run retrieval using each method over 100 queries and then compute the average time per query. Inference was run over a single Nvidia Quadro RTX 8000 GPU. We run each method with the maximum batch size that fits within the GPU. One parameter that highly affects the speed for both PathRetriever and MDR is the beam size. We use the default beam size for PathRetriever, which is 8, and we use a beam size of 5 for MDR, to closely match \model's pruning parameter $K_1 = 5$. Other than beam size, we use the default parameters for each method. 

Table~\ref{tab:inf-cost} shows the number of parameters of each method and the average time per query in seconds. First, we note that \model uses the most number of parameters since it is based on \txl while PathRetriever and MDR both rely on much smaller LMs such as BERT and RoBERTa. Interestingly, however, we can see that \model without ensembling has lower latency than PathRetriever, which is slowed down by the beam search process since it has to expand and encode outgoing links from each passage in the beam at each step. As expected, ensembling almost multiplies the latency of \model by the number of ensembles. 
Lastly, we note that MDR has significantly lower latency than both models, i.e., about 60x faster than PathRetriever and 46x than \model, which is mainly due to the fast implementation of the exact inner product search \cite{johnson2019billion}. It is worth noting, however, that MDR requires an expensive indexing step where \textit{every} document in the corpus (Wikipedia in our case) is encoded using the document encoder. \model, on the other hand, can work directly out-of-the-box without requiring such expensive indexing. 

\begin{table}[]
\footnotesize
    \centering
    \begin{tabular}{lcc}
    \toprule
    \textbf{System}     &   \textbf{\#params} & \textbf{Avg. query time(s)}\\
       \midrule
    PathRetriever & 110M & 1.95  \\
MDR & 125M &  0.03  \\
\midrule
\model & 3B & 1.38  \\ 
\model (ens) & 3B & 5.22 \\

\bottomrule
    \end{tabular}
    \cutcaptionup
    \caption{
    \footnotesize
    Inference cost and number of parameters of three systems comparing \model to PathRetriever and MDR. Query time is obtained by averaging the time to process 100 queries.
    }
    \cutcaptiondown
    \label{tab:inf-cost}
\end{table}

\subsection{Sensitivity to Document Order}
\label{app:doc-order}

Here, study \model's recall sensitivity to the document order in the prompt $\tau_c$ by running a simple experiment comparing two document ordering schemes:  \textbf{link-based} and \textbf{inverted link-based}. Link-based ordering is the standard approach used in \model, which orders the documents in the path based on their Wikipedia hyperlink traversal order. The inverted scheme, \textit{reverses} the order of the documents in the prompt. No instruction is used for this experiment.

\Cref{tab:doc-order} shows the retrieval performance with both orderings. Interestingly, reversing the order of the documents in the path does not seem to have a tangible effect on the reranking performance. While it is expected that $p(q | \tau_c)$ will change by reversing the document order in the prompt, it appears that the ranks of different paths remain almost unchanged, which explains why the recall is hardly affected. 

In other words, the path scores output by T5-XL does not appear to be sensitive to the document order prompt and can still. This might point to another benefit of LM-based path reranking: Since the performance is hardly affected by the document order, we do not have to worry about finding paths in the correct order (if such order exists) since the LM will still be able to assess the path relevance given different orders.

\begin{table}[t!]
\footnotesize
    \centering
    \begin{tabular}{ccccc}
    \toprule
    \textbf{Doc. ordering}     &   \textbf{R@2} & \textbf{R@10} & \textbf{AR@2} & \textbf{AR@10}\\
       \midrule
    \textit{\textbf{\tlarge}} &  &  &  & \\ 
    Link-based &  44.9 & 66.9 & 73.6 & 88.0 \\ 
    Inverted & 44.5 &  67.7 & 72.6 &   87.8 \\
    \midrule
    \textit{\textbf{\txl}} &  &  &  & \\ 
    Link-based &  44.6 & 67.9 & 74.1 & 88.2 \\ 
    Inverted & 45.7 &  69.0 & 74.4 &   88.3 \\
\bottomrule
    \end{tabular}
    \caption{Retrieval performance of \model using two different orderings of the documents in the prompt. Evaluation is done on a set of 2K examples from HotpotQA train set. \model exhibits minimal sensitivity to the document ordering.}
    \label{tab:doc-order}
\end{table}


\subsection{Comparison to Few-shot Systems}
\label{app:few-shot-compare}
So far, we have mainly compared \model to systems trained on many more examples. Here we compare \model to few-shot LOUVRE \cite{louvre2021} and PathRetriever \citep{Asai2019}. To this end, we train PathRetriever on $N$ examples from HotpotQA for $N \in {50, 100, 500, 1000}$ and compare its performance to \model ($N_{\text{demos}} = 10$). Since we were unable to obtain good performance by fine-tuning LOUVRE on few examples, we directly compare to the results reported in their paper, where 1\% of training data is used (\textasciitilde90 examples). \Cref{tab:fewshot-pathretriever} shows performance of both few-shot systems compared to \model. While PathRetriever's performance improves as we add more examples, we can see that it is much less data efficient than \model. Even with 1K examples i.e., around 10x more data than \model, it performs significantly worse across all metrics. We also observe that \model performs better than LOUVRE in terms of R@2 and AR@2 (more than 6 points better) and very close with respect to other metrics even though \model does\textit{ not} involve any (pre)training. 

\begin{table}[ht!]
\setlength{\tabcolsep}{2pt}
\footnotesize
    \centering
    \begin{tabular}{lccccc}
    \toprule
    \textbf{Approach}     & \textbf{\# Ex} &    \textbf{R@2} & \textbf{R@10} & \textbf{AR@2} & \textbf{AR@10}\\
       \midrule
    \multirow{4}{*}{PathRetriever} & 50 & 7.1 \tiny(4.4) & 14.5 \tiny(4.6) & 29.5 \tiny(6.2) & 40.0 \tiny(3.7) \\ 
     & 100 & 10.8 \tiny(1.1) & 19.1 \tiny(0.3) & 34.8 \tiny(1.5) & 43.1 \tiny(0.6) \\ 
     & 500 & 15.7 \tiny(0.3) & 22.4 \tiny(0.3) & 40.4 \tiny(0.3) & 46.4 \tiny(0.4) \\ 
     & 1K & 17.7 \tiny(0.4) & 23.6 \tiny(0.3) & 41.8 \tiny(0.3) & 47.1 \tiny(0.3) \\ 
    \midrule
        LOUVRE  & 1\%  &  53.5 \phantom{\tiny(0.0)} & 75.5\phantom{\tiny(0.0)} & 72.3\phantom{\tiny(0.0)}  & -- \\ 
        \midrule
    \model & \multirow{2}{*}{128} & \multirow{2}{*}{54.4} \phantom{\tiny(0.0)} & \multirow{2}{*}{73.5} \phantom{\tiny(0.0)}& \multirow{2}{*}{78.9} \phantom{\tiny(0.0)}& \multirow{2}{*}{88.9} \phantom{\tiny(0.0)}\\  
    ($N_{\text{demos}} = 10$) & & & & & \\
\bottomrule
    \end{tabular}
    \caption{Retrieval performance of \model compared to Few-shot PathRetriever. We show mean and (std) of PathRetriever's performance over 5 different seeds. The results of LOUVRE are take directly from \cite{louvre2021}. We observe that PathRetriever performs very poorly in low-data settings, even when using about 10x more data than \model. T
    }
    \label{tab:fewshot-pathretriever}
\end{table}
\end{document}